\documentclass[conference]{IEEEtran}
\usepackage{cite}
\usepackage{amsmath,amssymb,amsfonts}
\usepackage{algorithmic}
\usepackage{graphicx}
\usepackage{textcomp}
\usepackage{xcolor}
\usepackage{multirow}
\def\BibTeX{{\rm B\kern-.05em{\sc i\kern-.025em b}\kern-.08em
    T\kern-.1667em\lower.7ex\hbox{E}\kern-.125emX}}
    
\begin{document}

\title{Sentiment Analysis of Yelp Reviews: \\ A Comparison of Techniques and Models}

\author{
\IEEEauthorblockN{Siqi Liu}
\IEEEauthorblockA{\textit{University of Waterloo}\\
sq2liu@uwaterloo.ca}
}

\maketitle

\begin{abstract}
We use over 350,000 Yelp reviews on 5,000 restaurants to perform an ablation study on text preprocessing techniques. We also compare the effectiveness of several machine learning and deep learning models on predicting user sentiment (negative, neutral, or positive). For machine learning models, we find that using binary bag-of-word representation, adding bi-grams, imposing minimum frequency constraints and normalizing texts have positive effects on model performance. For deep learning models, we find that using pre-trained word embeddings and capping maximum length often boost model performance. Finally, using macro F1 score as our comparison metric, we find simpler models such as Logistic Regression and Support Vector Machine to be more effective at predicting sentiments than more complex models such as Gradient Boosting, LSTM and BERT.

\end{abstract}


\section{Introduction}

\subsection{Background}

Sentiment analysis is the process of analyzing a piece of text and determining its author's attitude. Using modern natural language processing (NLP) techniques, one can predict the most likely emotional tone that the author carries based on what's presented in the text. Sentiment analysis is widely used by companies around the world to evaluate customer satisfaction, conduct market research and monitor brand reputation.

Yelp is a popular crowd-sourced review platform with millions of active users who rate and review hundreds of thousands of businesses across the globe. Since 2016, Yelp has been releasing and updating subsets of its enormous database to the general public.

\subsection{Goal}

While there is no one-fits-all approach to performing sentiment analysis, the goal of this project is to provide some guidance for practitioners to consider when developing models that fit their needs.

There are two main steps to building a model: data preparation and model selection. First, we perform an ablation study on several text preprocessing techniques (e.g., stop word removal, normalization) using a simple multinomial Naive Bayes model \cite{dasilva2014}. Afterwards, using the preprocessed data, we train several machine learning (e.g., Logistic Regression, Support Vector Machine) and deep learning (e.g., LSTM, BERT) models and compare their effectiveness at predicting sentiments.

\section{Data}

\subsection{Preparation}

We download our data from the Yelp Open Data website (https://www.yelp.com/dataset). For the scope of this project, we only use the information on reviews, ratings and businesses, and forgo other information such as tips, pictures and user data that could also help predict user sentiment.

The raw dataset contains 6.7 million reviews and ratings on 192,609 businesses. Due to limited computational power, we narrow the scope of our study to businesses that are:

\begin{itemize}
    \item In the restaurant/food servicing industry
    \item Located in the Greater Toronto Area
    \item With at least 10 reviews
\end{itemize}

\begin{figure}[htbp]
\centerline{
\includegraphics[width=\columnwidth]{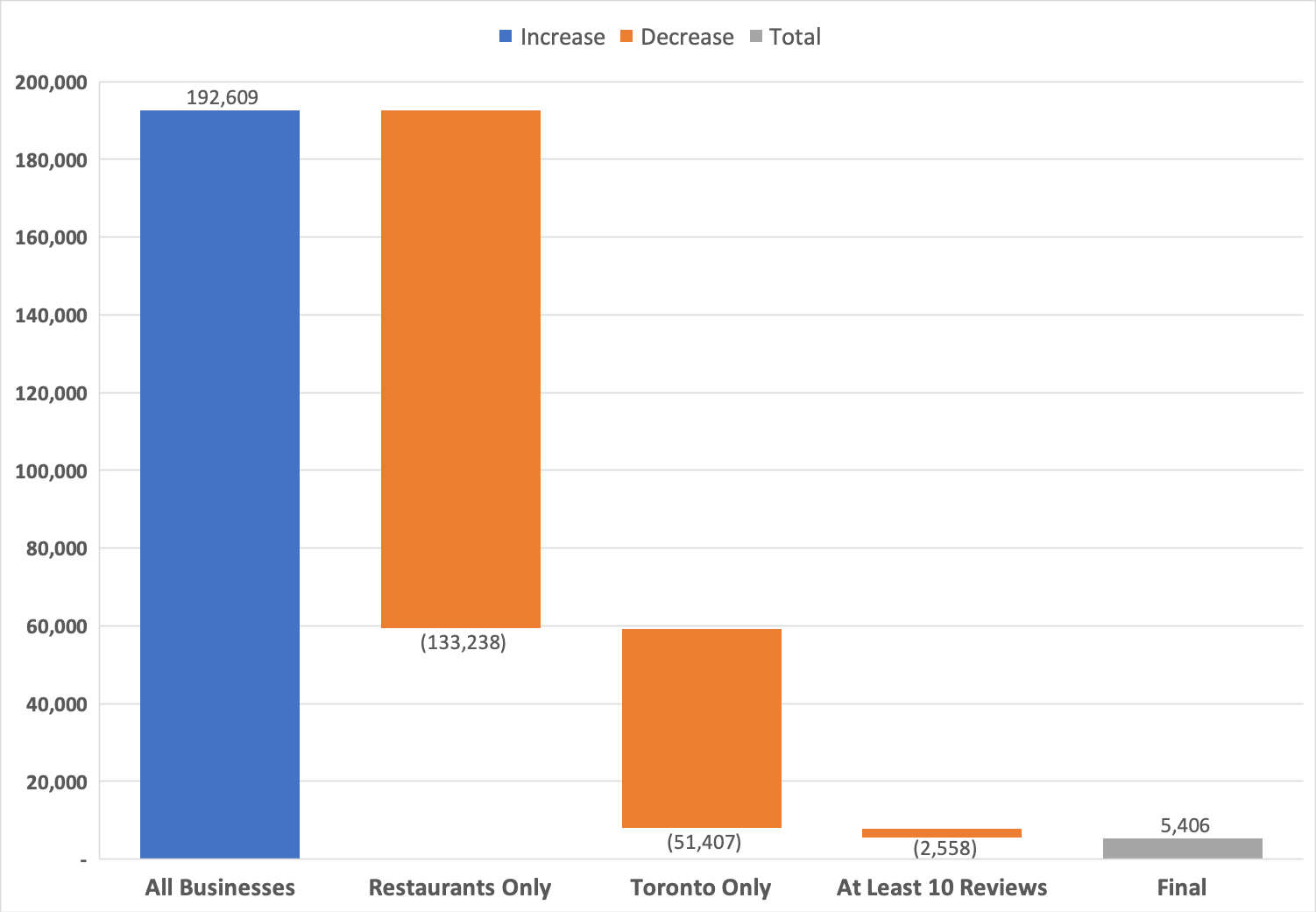}
}
\caption{Study Population Selection Waterfall}
\label{fig}
\end{figure}

Fig. 1 shows the impacts of each of our narrowing criteria. As a result, our final study population consists of 5,406 businesses. We then select all 362,554 reviews and ratings on these businesses for our dataset.

There are two columns in our dataset - \textit{review} contains the text reviews, and \textit{rating} contains the star ratings that accompany the reviews. The ratings range from 1 to 5, with 1 being the lowest and 5 being the highest. This is a multi-class classification problem, and our goal is to build classification models that can accurately predict the ratings based on the reviews.

\subsection{Distribution}

Before jumping into model building, it is important to understand the distribution of our target variable (i.e., ratings), as having an imbalanced dataset could lead to sub-optimal model performance.

\begin{figure}[htbp]
\centerline{
\includegraphics[width=\columnwidth]{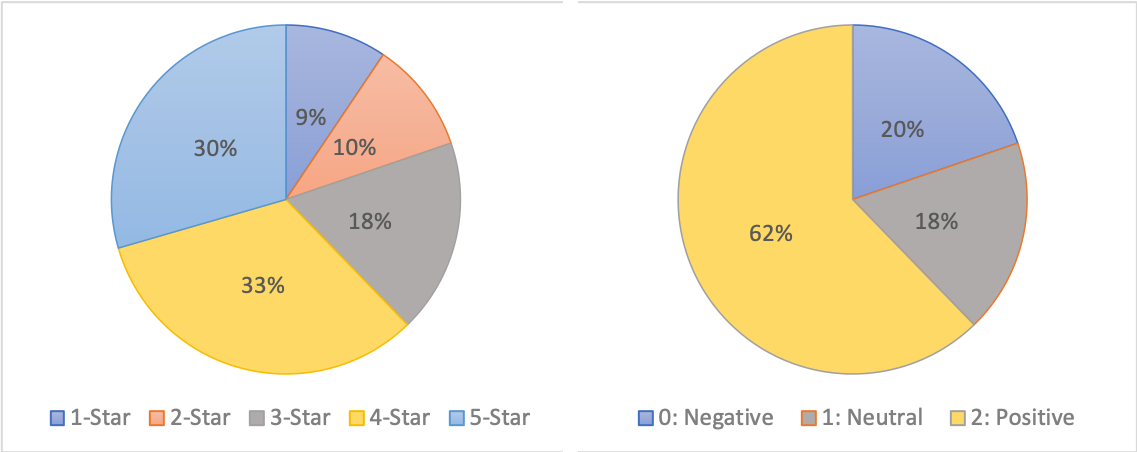}
}
\caption{Target Class Ratio - Before (left) and After (right) Grouping}
\label{fig}
\end{figure}

Fig. 2 (left) shows the proportion of ratings in our dataset. It turns out that only 9\% of our dataset has 1-star ratings, compared to 4-star ratings and 5-star ratings, which occupy a third of our dataset each. From an analytical perspective, businesses like to know more about the difference between a 1-star rating and a 5-star rating, rather than the difference between a 1-star rating and a 2-star rating. Therefore, we group ratings 1 and 2 as ``negative`` sentiments, group ratings 4 and 5 as ``positive`` sentiments, and consider rating 3 as ``neutral``. Fig. 2 (right) shows the target class ratio after grouping. For modelling purposes, we use 0 for negative sentiments, 1 for neutral, and 2 for positive sentiments. Later on, we will examine the effect on model performance when the dataset is completely balanced.

\subsection{Train Test Split}

Since we will be comparing model performances, we need to establish a universal test set. Using \textit{scikit-learn}, we split the dataset into 75\% training and 25\% testing, while preserving the target class ratio. In the end, we have 271,915 data points for training and 90,639 data points for testing.

\section{Evaluation Metric}

We will use macro F1 score defined in Sokolova, 2009 \cite{sokolova2009} as the evaluation metric for measuring and comparing model performances:

\begin{equation}
\text{Precision}_{M} = \frac{\sum_{i = 1}^{l}\frac{tp_{i}}{tp_{i} + fp_{i}}}{l}\label{eq}
\end{equation}

\begin{equation}
\text{Recall}_{M} = \frac{\sum_{i = 1}^{l}\frac{tp_{i}}{tp_{i} + fn_{i}}}{l}\label{eq}
\end{equation}

\begin{equation}
\text{F1}_{M} = \frac{2 \times \text{Precision}_{M} \times \text{Recall}_{M}}{\text{Precision}_{M} + \text{Recall}_{M}}
\end{equation}

where $l$ is the total number of classes and $tp$, $fp$, $fn$ are the number of True Positives, False Positives and False Negatives respectively.

\begin{table}[htbp]
\caption{Example Confusion Matrix}
\begin{center}
\begin{tabular}{ccccc}
 &  & \multicolumn{3}{c}{Predicted} \\ \cline{3-5} 
 & \multicolumn{1}{c|}{} & \multicolumn{1}{c|}{Class 0} & \multicolumn{1}{c|}{Class 1} & \multicolumn{1}{c|}{Class 2} \\ \cline{2-5} 
\multicolumn{1}{c|}{\multirow{3}{*}{Actual}} & \multicolumn{1}{c|}{Class 0} & \multicolumn{1}{c|}{1} & \multicolumn{1}{c|}{0} & \multicolumn{1}{c|}{1} \\ \cline{2-5} 
\multicolumn{1}{c|}{} & \multicolumn{1}{c|}{Class 1} & \multicolumn{1}{c|}{0} & \multicolumn{1}{c|}{1} & \multicolumn{1}{c|}{1} \\ \cline{2-5} 
\multicolumn{1}{c|}{} & \multicolumn{1}{c|}{Class 2} & \multicolumn{1}{c|}{1} & \multicolumn{1}{c|}{0} & \multicolumn{1}{c|}{3} \\ \cline{2-5} 
\end{tabular}
\label{tab1}
\end{center}
\end{table}

Consider the example in Table I. In this example, macro precision, macro recall and macro F1 scores are computed as follows:

\begin{equation}
\text{Precision}_{M} = \frac{\frac{1}{1+1}+\frac{1}{1+0}+\frac{3}{3+2}}{3} = \frac{0.5+1.0+0.6}{3} = 0.7 \label{eq}
\end{equation}

\begin{equation}
\text{Recall}_{M} = \frac{\frac{1}{1+1}+\frac{1}{1+1}+\frac{3}{3+1}}{3} = \frac{0.5+0.5+0.75}{3} = 0.5833 \label{eq}
\end{equation}

\begin{equation}
\text{F1}_{M} = \frac{2 \times 0.7 \times 0.5833}{0.7 + 0.5833} = 0.63
\end{equation}

Note that this macro F1 score is slightly different from the macro F1 score implemented in \textit{scikit-learn}, which is a simple average of within-class F1 scores.

\section{Ablation Study Results}

\subsection{Training Set Size}

We want to examine the marginal improvements in the model performance of having a larger training set. Many studies (e.g., Renault, 2019 \cite{renault2019}) have shown the importance of having a large dataset on sentiment analysis tasks. Starting with a small training set size of 1,000, we train eight multinomial Naive Bayes models using default parameters and record the macro F1 scores.

\begin{figure}[htbp]
\centerline{
\includegraphics[width=\columnwidth]{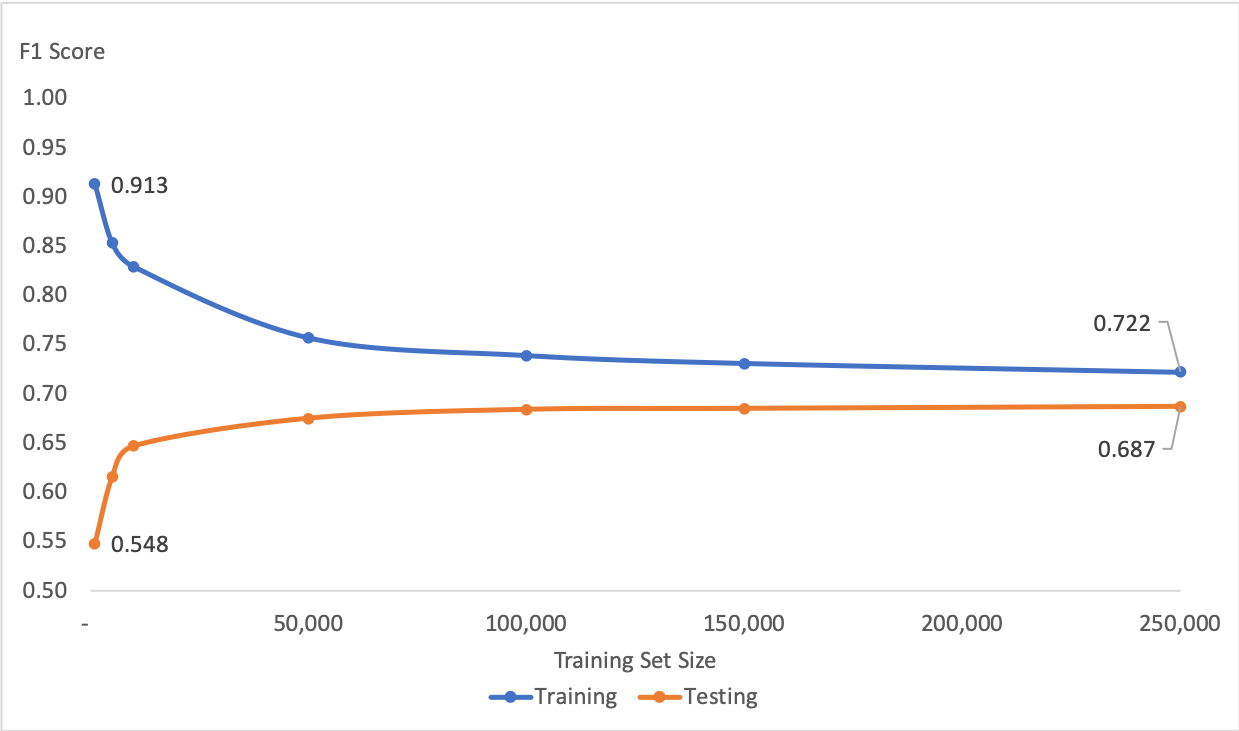}
}
\caption{F1 Score by Training Set Size}
\label{fig}
\end{figure}

\begin{figure}[htbp]
\centerline{
\includegraphics[width=\columnwidth]{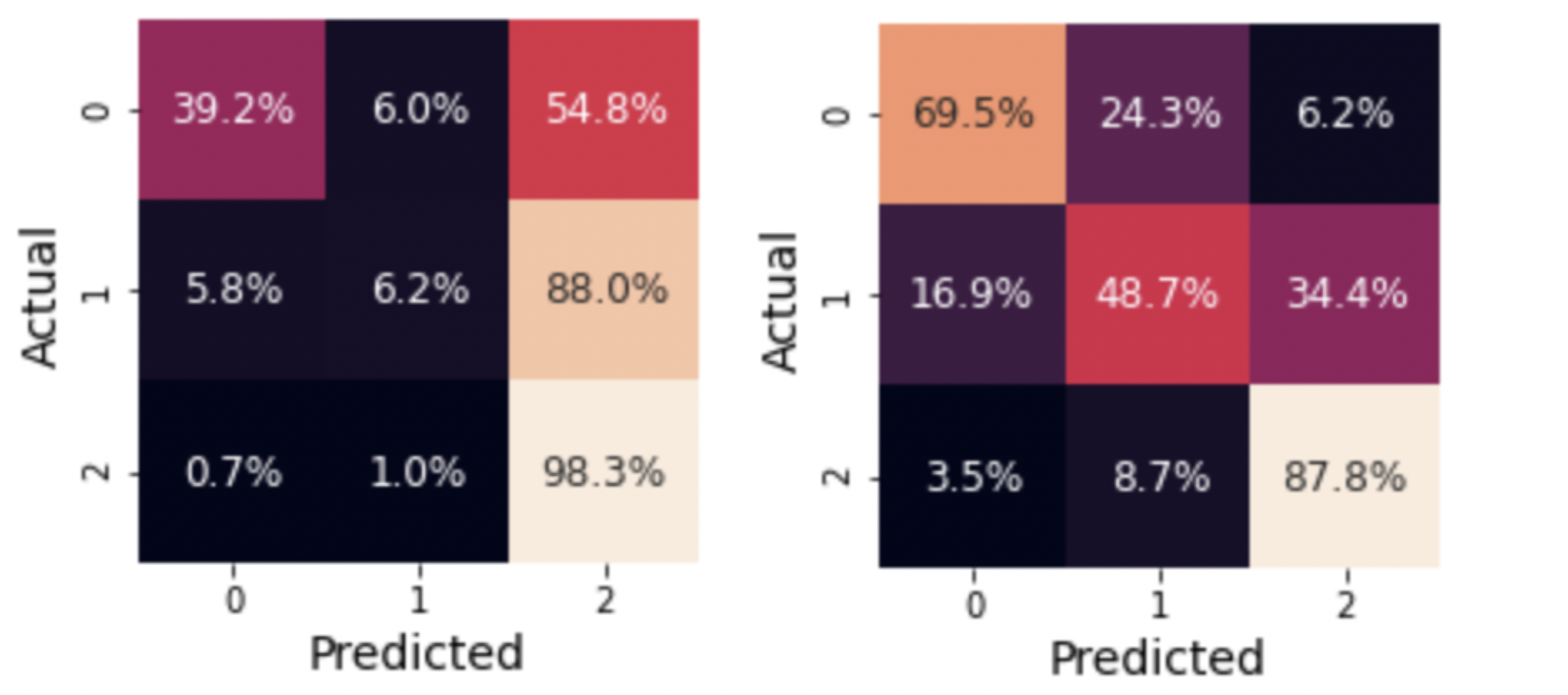}
}
\caption{Size = 1,000 (left) vs. Size = 25,000 (right)}
\label{fig}
\end{figure}

Fig. 3 shows the model performance on the training set and the test set. Fig. 4 shows the normalized confusion matrices on then test set when training set sizes are 1,000 (left) and 250,000 (right). As we can see, although the marginal benefit of having a large dataset diminishes over time, it prevents overfitting effectively and improves overall model performance significantly.

\subsection{Balanced Target Class Ratio}
Currently, our target class ratio is at 20/20/60. While this is not considered severely imbalanced, we want to examine if having a completely balanced dataset could improve model performance.

Since our smallest class has a training set size of 49,000, we set up our balanced training set by down-sampling all three classes to 45,000 data points each. For our benchmark case, we down-sample the entire training set to 45,000 * 3 = 135,000 data points, preserving the 20/20/60 imbalanced target class ratio.

\begin{figure}[htbp]
\centerline{
\includegraphics[width=\columnwidth]{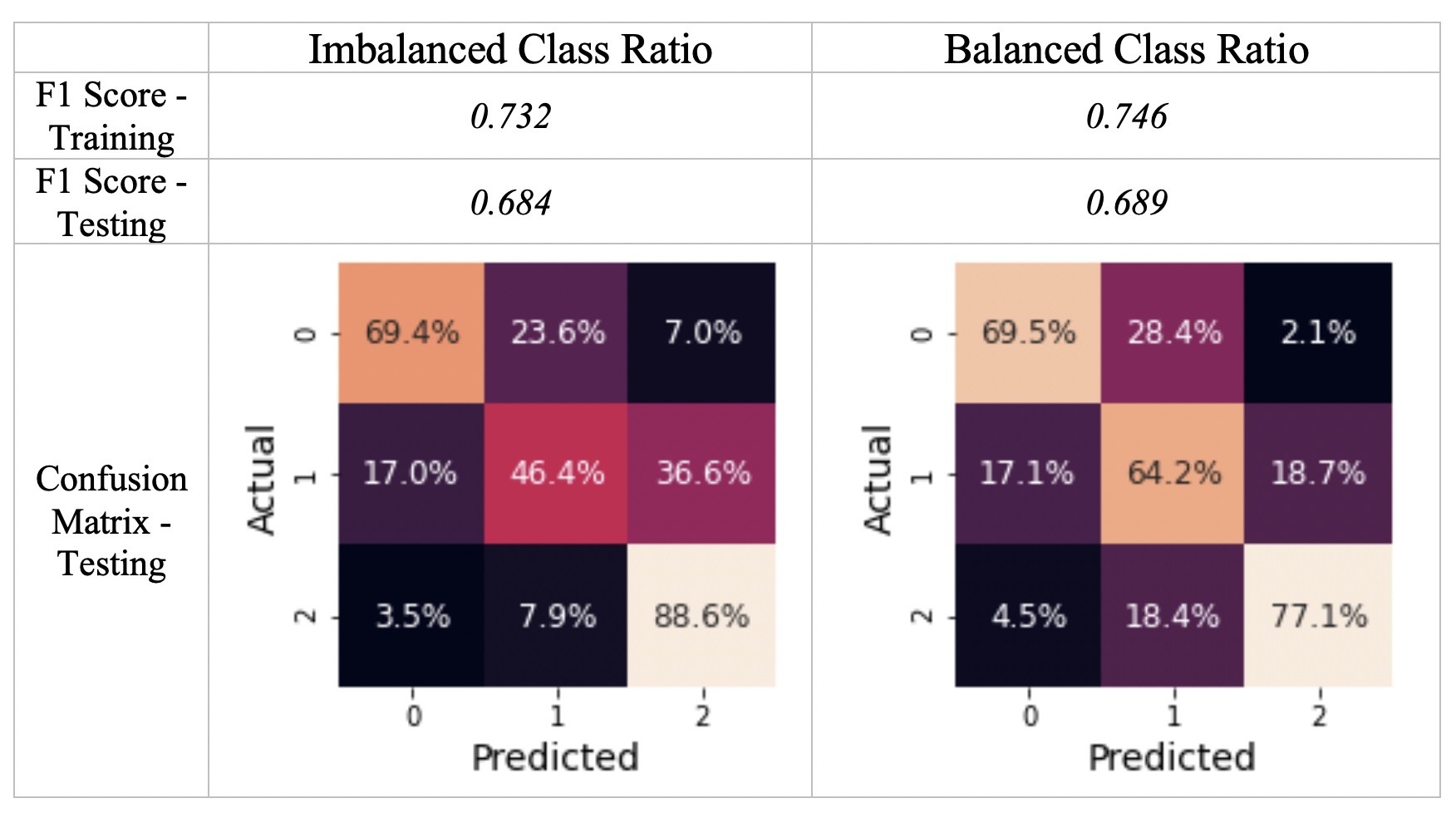}
}
\caption{Imbalanced (left) vs. Balanced (right)}
\label{fig}
\end{figure}

Fig. 5 shows the results when models are trained with imbalanced (left) and balanced (right) training sets. As we can see, having a balanced training set not only improves the model performance, it also prevents the model from always predicting the majority class (e.g., actual = 1, predicted = 2 goes down from 36.6\% to 18.7\%). Although the True Positive rate for class 2 decreases from 88.6\% to 77.1\%, this is a reasonable trade-off for lower False Negative rates in the other two classes.

Moreover, further experiments show that the model trained on the down-sampled balanced dataset outperforms the model trained on the entire training set without any down-sampling, despite the difference in sample size. Therefore, we update our training set for all future ablation studies and model builds to be the down-sampled balanced training set. We do not, however, modify the test set, since the class imbalanceness is an actual representation of the distribution in our target population.

\subsection{Word Representation}

Now that we have our final dataset, we can start building our text preprocessing pipeline. First, for each text corpus, we need to extract features for our models to use. Two of the most common representation methods are bag-of-words and term frequency-inverse document frequency (TF-IDF).

Under the bag-of-words representation, each text corpus is converted into a vector of word counts, so the feature space becomes all unique words among all training texts. One variation of this model is to use binary indicators rather than word counts. The motivation behind this variation is that sometimes, word occurrences are more important than word frequency.

The TF-IDF representation also converts each text corpus into a vector. However, instead of word counts, it is

\begin{equation}
\text{tf-idf}(t, d, D) = \text{tf}(t, d) \times \text{idf}(t, D) \label{eq}
\end{equation}

where $\text{tf}(t, d)$ is the term frequency of the term $t$ in document $d$, and $\text{idf}(t, D)$ is the inverse (log) document frequency of $t$ in the entire training set of documents $D$. $\text{tf-idf}(t, d, D)$ is high if the term has high term frequency and low document frequency. Essentially, this representation emphasizes terms that are rare and discounts terms that are less valuable because they appear in a lot of documents.

\begin{figure}[htbp]
\centerline{
\includegraphics[width=\columnwidth]{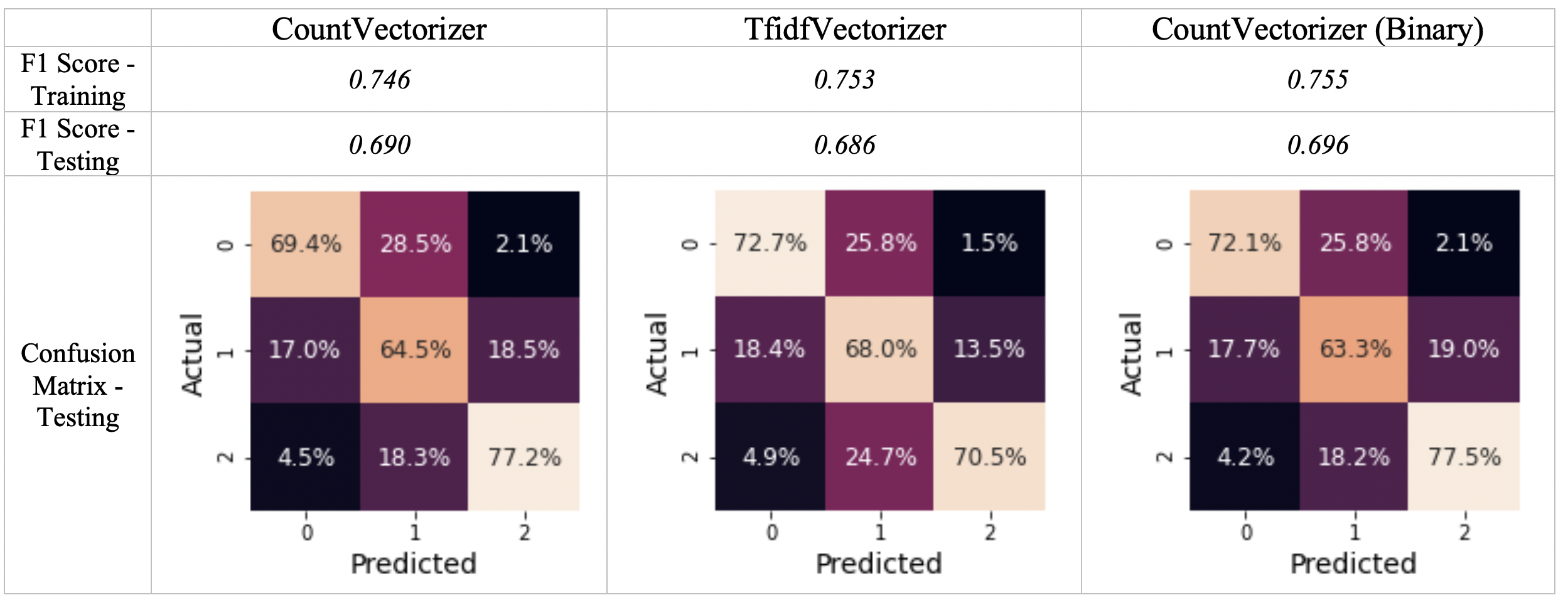}
}
\caption{CountVectorizer (left) vs. TfidfVectorizer (middle) vs. Binary CountVectorizer (right)}
\label{fig}
\end{figure}

Fig. 6 shows the results when models are trained using each of the three representation methods through \textit{CountVectorizer} and \textit{TfidfVectorizer} in \textit{scikit-learn}. Even though the \textit{TfidfVectorizer} model has the lowest F1 score, it has higher True Positive rates for both class 0 and class 1 compared to both \textit{CountVectorizer} models. Since the binary variation of \textit{CountVectorizer} achieves the highest F1 score, we will use it for our future experiments.

\subsection{Number of N-Grams}

Let us now investigate the benefit of incorporating n-gram into our model. N-gram is simply a contiguous sequence of n words. For example, the text corpus ``the food is not good`` has:

\begin{itemize}
    \item Five 1-gram/unigram words: ``the``, ``food``, ``is``, ``not``, ``good``
    \item Four 2-gram/bigram phrases: ``the food``, ``food is``, ``is not``, ``not good``
\end{itemize}

Adding bigram, trigram or even four-gram phrases into our feature space allows us to capture modified verbs and nouns, thus improve model performance (Wang, 2012 \cite{wang2012}).

\begin{figure}[htbp]
\centerline{
\includegraphics[width=\columnwidth]{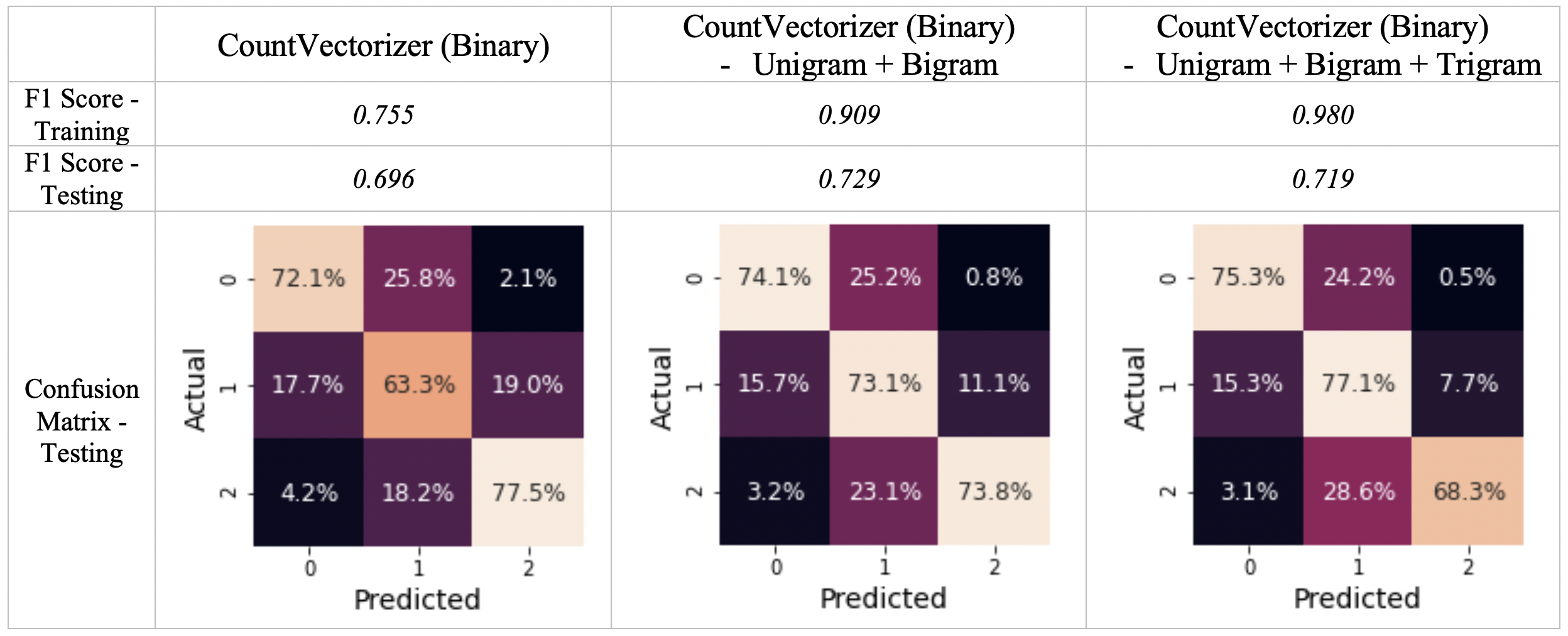}
}
\caption{Unigram (left) vs. Unigram + Bigram (middle) vs. Unigram + Bigram + Trigram (right)}
\label{fig}
\end{figure}

For our experiment, we compare three n-gram models - unigram, bigram and trigram. Fig. 7 shows the results of our experiment. As we include more n-grams, the model tends to overfit since the vocabulary size increases. While the True Positive rates for class 0 and 1 both increase, it decreases for class 2. Consider one particular text corpus from the test set:

\begin{quote}
\textit{Food wasn't great. My gnocchi was incredibly greasy. Would not recommend anyone coming here.}
\end{quote}

This review has a ground-truth label of 0. While it is correctly predicted by our bigram model, our unigram model thinks that it belongs to class 2.

\begin{table}[htbp]
\caption{Conditional Log-Likelihoods}
\begin{center}
\resizebox{\columnwidth}{!}{
\begin{tabular}{c|c|c|c|c|c|c|}
\cline{2-7}
 & \multicolumn{3}{c|}{Unigram} & \multicolumn{3}{c|}{Unigram + Bigram} \\ \cline{2-7} 
 & Class 0 & Class 1 & Class 2 & Class 0 & Class 1 & Class 2 \\ \hline
\multicolumn{1}{|c|}{food} & -4.878 & -5.023 & -5.019 & -6.127 & -6.184 & -6.133 \\
\multicolumn{1}{|c|}{greasy} & -8.485 & -8.471 & -8.792 & -9.246 & -9.244 & -9.616 \\
\multicolumn{1}{|c|}{great} & -6.531 & -5.934 & -5.279 & -7.404 & -6.880 & -6.351 \\
\multicolumn{1}{|c|}{incredibly} & -8.987 & -9.464 & -8.933 & -9.722 & -10.191 & -9.714 \\
\multicolumn{1}{|c|}{incredibly greasy} &  &  &  & -13.790 & -14.882 & -16.118 \\
\multicolumn{1}{|c|}{not} & -4.665 & -4.862 & -5.296 & -6.038 & -6.139 & -6.473 \\
\multicolumn{1}{|c|}{not recommend} &  &  &  & -9.269 & -10.747 & -11.761 \\
\multicolumn{1}{|c|}{recommend} & -7.536 & -7.561 & -6.719 & -8.231 & -8.314 & -7.516 \\
\multicolumn{1}{|c|}{would} & -5.748 & -5.711 & -5.921 & -6.794 & -6.734 & -6.919 \\
\multicolumn{1}{|c|}{would not} &  &  &  & -8.666 & -9.404 & 10.417 \\ \hline
\multicolumn{1}{|c|}{Predicted Prob} & 0.343 & 0.129 & 0.528 & 0.999 & 0.001 & 0.000 \\ \hline
\end{tabular}}
\label{tab2}
\end{center}
\end{table}

Table II shows the conditional log-likelihoods ($\log{\text{Pr}(x|\text{class})}$) of some of the features predicted by our unigram and bigram models. We can see that:

\begin{itemize}
    \item For any given word (e.g., \textit{food}, \textit{great}), the conditional log-likelihood across all three classes is higher (less negative) in the unigram model than in the bigram model. This is because the vocabulary size is larger in the bigram model, resulting in lower relative-frequency for each word
    \item The ``trend`` for conditional log-likelihoods among the three classes is the same in the unigram model and the bigram model. For example, under both models, the word \textit{incredibly} has higher log-likelihoods for classes 0 and 2 than class 1. This can be interpreted as: if someone expresses strong feelings in his/her review, he/she is more likely to be either negative or positive than neutral
    \item Under the bigram model, phrases like \textit{incredibly greasy}, \textit{not recommend}, \textit{would not} are captured. All three phrases assign a much higher log-likelihood to class 0 than to class 2. Even though their subset words (e.g., \textit{recommend}, \textit{would}) favour class 2 over class 0, the bigram phrases' high log-likelihoods more than offset these favours and tilt the prediction in the bigram model to a 99\% probability toward class 0
\end{itemize}

This particular example illustrates the importance of including contiguous sequences of words in our model. However, trigram model suffers severely from overfitting, as the vocabulary size grows too large.

\subsection{Remove Stopwords}

Stopwords are commonly used words, such as ``the``, ``and``, ``to``. They appear in almost every text corpus, so naturally, we would like to remove them from our models.

\begin{figure}[htbp]
\centerline{
\includegraphics[width=\columnwidth]{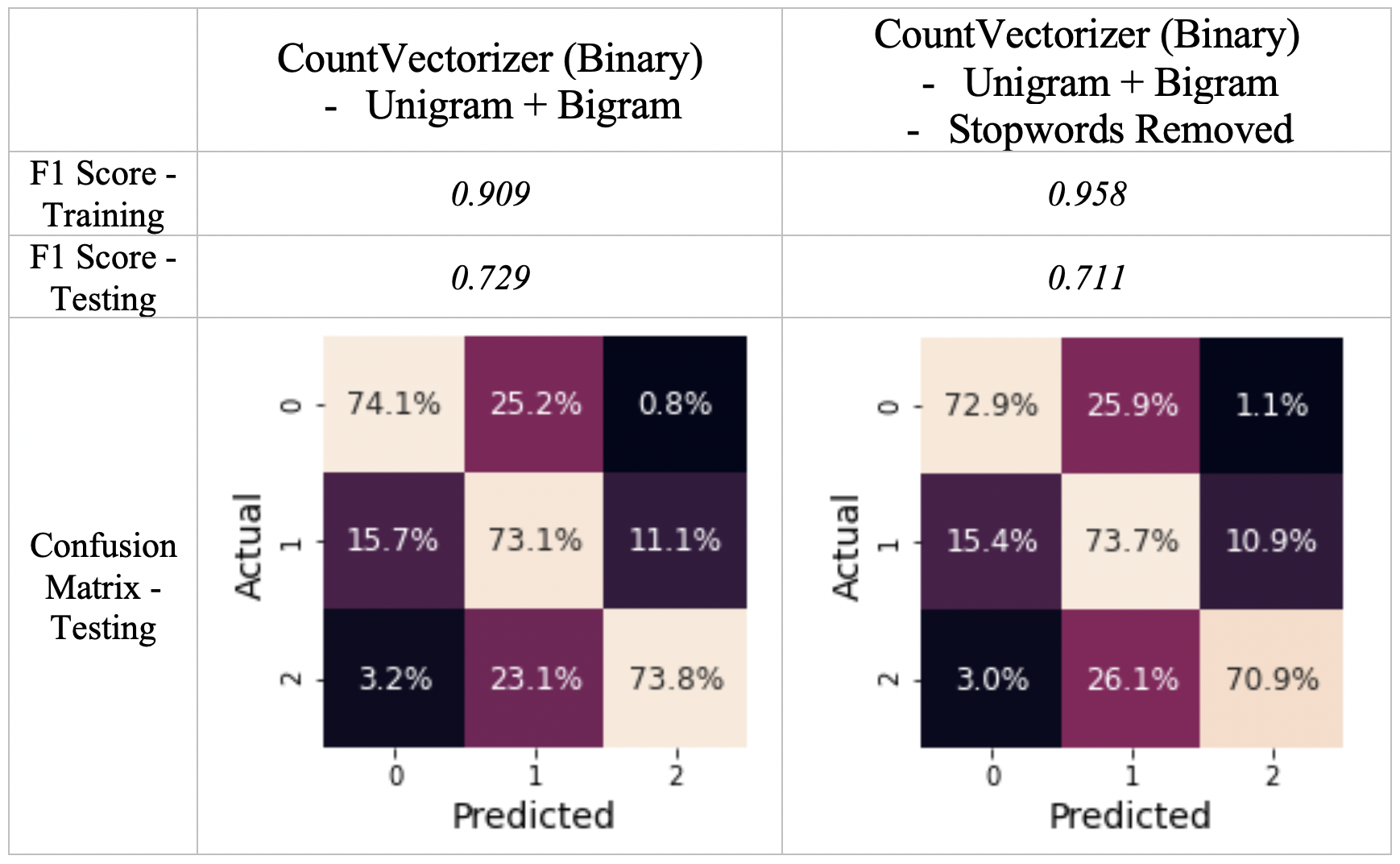}
}
\caption{Stopwords Not Removed (left) vs. Removed (right)}
\label{fig}
\end{figure}

\begin{table}[htbp]
\caption{Top-10 Most Frequent Words}
\begin{center}
\begin{tabular}{c|c|c|}
\cline{2-3}
 & \begin{tabular}[c]{@{}c@{}}Not\\ Removed\end{tabular} & Removed \\ \hline
\multicolumn{1}{|c|}{1} & the & food \\ \hline
\multicolumn{1}{|c|}{2} & and & good \\ \hline
\multicolumn{1}{|c|}{3} & to & place \\ \hline
\multicolumn{1}{|c|}{4} & it & service \\ \hline
\multicolumn{1}{|c|}{5} & of & like \\ \hline
\multicolumn{1}{|c|}{6} & was & would \\ \hline
\multicolumn{1}{|c|}{7} & for & one \\ \hline
\multicolumn{1}{|c|}{8} & is & back \\ \hline
\multicolumn{1}{|c|}{9} & but & really \\ \hline
\multicolumn{1}{|c|}{10} & in & great \\ \hline
\end{tabular}
\end{center}
\end{table}

\textit{NLTK}, a popular natural language toolkit, provides a list of 179 most common stopwords. Fig. 8 shows the results before (left) and after (right) we remove these stopwords from our models. Table III shows the resulting top-10 most frequent words. Interestingly, the model performance decreases after we remove these stopwords. This corroborates with the findings in Saif, 2014 \cite{saif2014}, which shows that using a pre-compiled list of stop words negatively impacts the performance of sentiment classification.

\begin{figure}[htbp]
\centerline{
\includegraphics[width=\columnwidth]{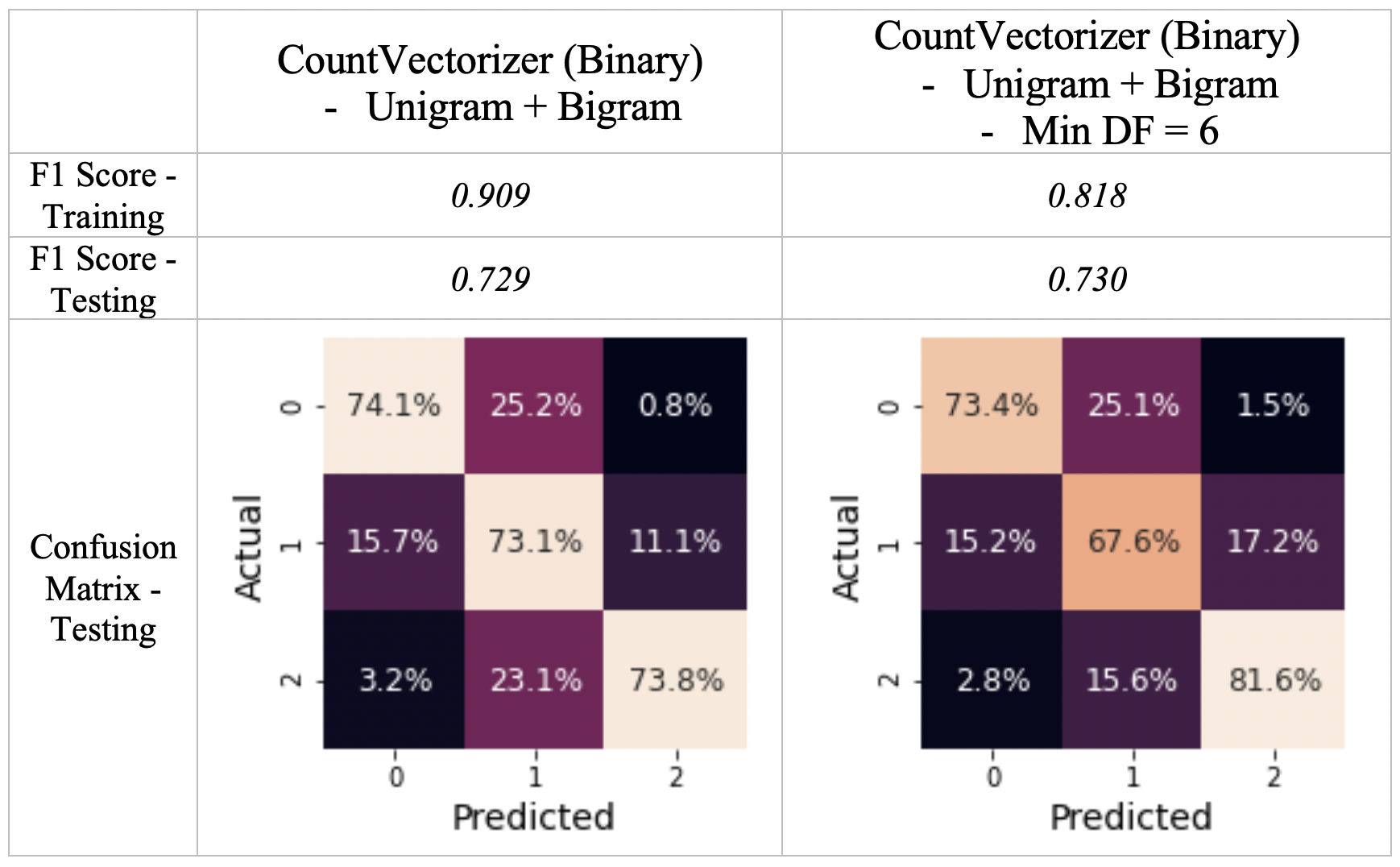}
}
\caption{Without (left) vs. With (right) Minimum Frequency Requirement}
\label{fig}
\end{figure}

Instead, Saif, 2014 \cite{saif2014} suggests to eliminate infrequent terms. As such, we set up another experiment, in which we eliminate all words and phrases that appear in at most 5 reviews. Fig. 9 shows the result before (left) and after (right) we impose the minimum frequency requirement on our bigram model. Although the True Positive rates for class 0 and 1 decrease, there is a significant improvement on overfitting and the overall F1 score. More importantly, the feature space shrinks by over 85\% from 2,130,891 to 264,977, resulting in much faster training and testing time.

\subsection{Normalization}

Normalization is a powerful and effective way to scale down feature space by reducing words to their roots. We will experiment with two popular normalization techniques - stemming and lemmatization.

\begin{table}[htbp]
\caption{Example Text Normalization}
\begin{center}
\begin{tabular}{c|c|c|c|}
\cline{2-4}
\multicolumn{1}{l|}{} & Original & Stemmed & Lemmatized \\ \hline
\multicolumn{1}{|c|}{1} & what a trouble & what a troubl & what a trouble \\ \hline
\multicolumn{1}{|c|}{2} & this is very troubling & thi is veri troubl & this be very troubling \\ \hline
\multicolumn{1}{|c|}{3} & i am troubled & i am troubl & i be trouble \\ \hline
\end{tabular}
\end{center}
\end{table}

Table IV shows three sentences after passing through \textit{PorterStemmer} and \textit{WordNetLemmatizer} in \textit{NLTK}. With stemming, ``trouble'', ``troubling'' and ``troubled'' all get reduced to ``troubl'', which is not a word. This is common in stemming. On the other hand, with lemmatization, the results are still words, and ``troubling'', which is an adjective, remains to be ``troubling'' instead of ``trouble''. This is because, with lemmatization, we can also do part-of-speech tagging, which determines the identification of a word (e.g., noun, verb, adjective) depending on the context. However, this system is not perfect, as ``troubled'' is also an adjective in this context but still gets reduced to ``trouble''. Overall, lemmatization with part-of-speech tagging attempts to minimize the loss in the original meaning, at the expense of lower feature space reduction than stemming.

\begin{figure}[htbp]
\centerline{
\includegraphics[width=\columnwidth]{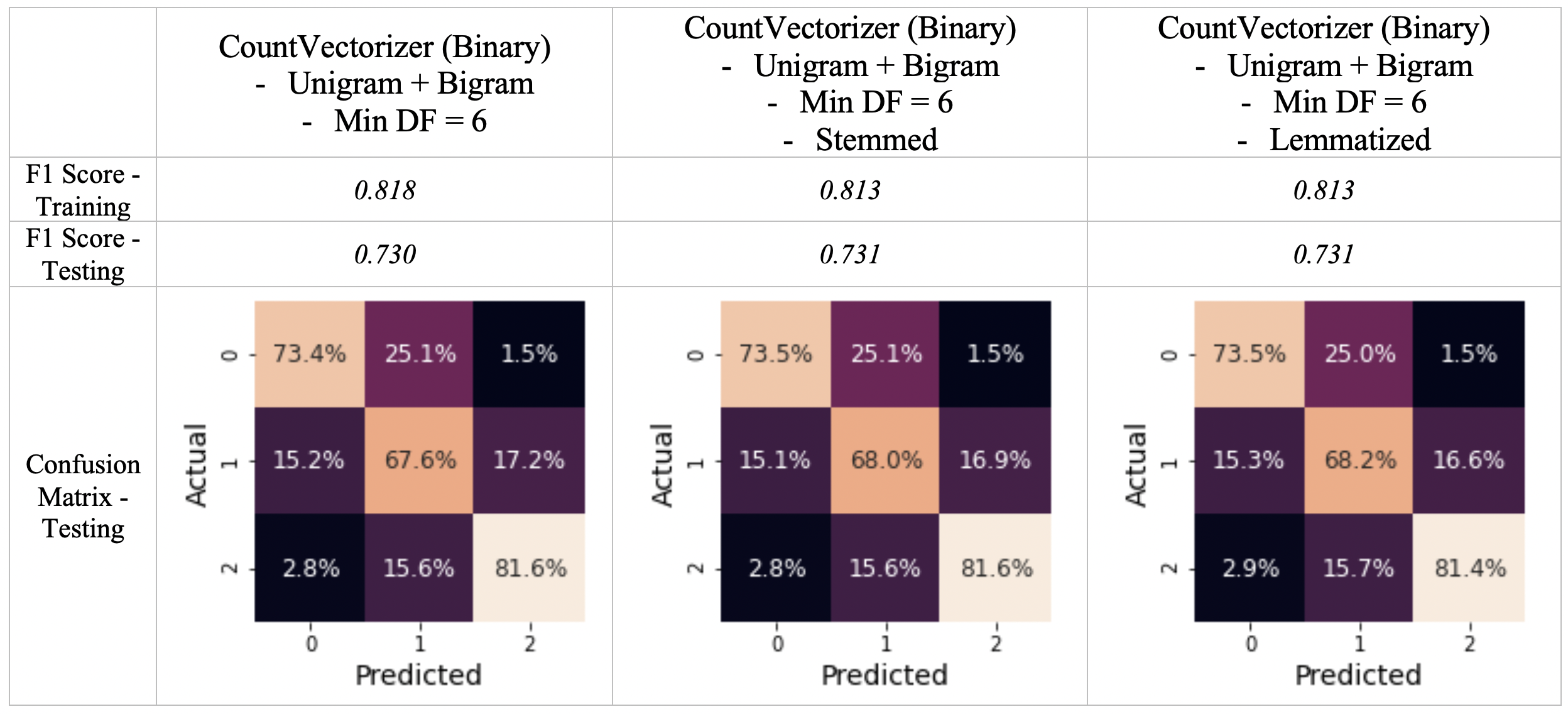}
}
\caption{Original (left) vs. Stemmed (middle) vs. Lemmatized (right)}
\label{fig}
\end{figure}

Fig. 10 shows the results after we perform text normalization. As we can see, there is very little improvement in model performance. However, considering the reduction in overfitting and feature space, we will continue to lemmatize our texts for our model builds.

Finally, our text preprocessing pipeline is as follows:

\begin{itemize}
    \item Use the binary variation of \textit{CountVectorizer}
    \item Unigram + bigram
    \item Minimum document frequency of 6
    \item Lemmatized, with part-of-speech tagging
\end{itemize}

When trained on a multinomial Naive Bayes model, we obtain a macro F1 score of 0.731 on the test set. In the next section, we will experiment with other types of models to see if we can outperform this score.

\section{Model Results}

We consider three popular machine learning models for text classification - Logistic Regression \cite{hosmer2013}, Support Vector Machine (SVM) \cite{suykens1999} and Gradient Boosting (XGBoost) \cite{chen2016}. We also consider two deep learning architectures that are widely used for NLP tasks - Long Short Term Memory (LSTM) \cite{hochreiter1997} and Bidirectional Encoder Representations (BERT) \cite{devlin2018}.

\begin{table}[htbp]
\caption{Model Results}
\begin{center}
\begin{tabular}{c|c|c|}
\cline{2-3}
 & Macro F1 Score (Test) & Training Time \\ \hline
\multicolumn{1}{|c|}{Multinomial Naive Bayes} & 0.731 & 341 ms \\ \hline
\multicolumn{1}{|c|}{Logistic Regression} & 0.757 & 2 min 5 sec \\ \hline
\multicolumn{1}{|c|}{Support Vector Machine} & 0.757 & 11 sec \\ \hline
\multicolumn{1}{|c|}{Gradient Boosting (XGBoost)} & 0.750 & 1 hr 25 min  \\ \hline
\multicolumn{1}{|c|}{LSTM} & 0.701 & 18 min 20 sec \\ \hline
\multicolumn{1}{|c|}{BERT} & 0.723 & 3 hr 15 min  \\ \hline
\end{tabular}
\end{center}
\end{table}

Table V shows the model performances, as well as the training times for all of our trained models. All models are trained on Google Colab (some using GPU/TPU backends). Note that training time does not account for the time for hyperparameter tuning, which could take hours or even days depending on the number of hyperparameters. As it turns out, more complex models (Gradient Boosting, LSTM, BERT) do not necessarily improve model performance. This corroborates with the findings in many other papers related to sentiment analysis in social media (e.g., Salinca, 2015 \cite{salinca2015} and Xu, 2014 \cite{xu2014}).

\subsection{Logistic Regression, SVM}

Logistic Regression models offer great interpretability and transparency, something that other machine learning and deep learning models often lack. Coefficients in a logic regression model are log odds. Since our problem is a multi-class one, each feature has three coefficients (one for each class).

\begin{table}[htbp]
\caption{Most Negative (left), Neutral (middle) and Positive (right) Words/Phrases}
\begin{center}
\begin{tabular}{c|c|c|c|}
\cline{2-4}
 & Negative & Neutral & Positive \\ \hline
\multicolumn{1}{|c|}{1} & horrible & 3 star & delicious \\ \hline
\multicolumn{1}{|c|}{2} & terrible & 3 5 & excellent \\ \hline
\multicolumn{1}{|c|}{3} & bland & three star & amazing \\ \hline
\multicolumn{1}{|c|}{4} & disappointing & ok & awesome \\ \hline
\multicolumn{1}{|c|}{5} & 2 star & a ok & perfect \\ \hline
\multicolumn{1}{|c|}{6} & disappointed & average & fantastic \\ \hline
\multicolumn{1}{|c|}{7} & not worth & decent & amaze \\ \hline
\multicolumn{1}{|c|}{8} & rude & not bad & best \\ \hline
\multicolumn{1}{|c|}{9} & poor & okay & 4 star \\ \hline
\multicolumn{1}{|c|}{10} & awful & however & incredible \\ \hline
\end{tabular}
\end{center}
\end{table}

Table VI shows the most negative, neutral and positive words and phrases based on our trained logistic regression model. For example, the most negatively word is ``horrible'', with fitted coefficients of (0.593, -0.11, -0.483) for classes 0, 1 and 2 respectively.

\begin{table}[htbp]
\caption{Most (left) vs. Least (right) Discriminative Words/Phrases}
\begin{center}
\begin{tabular}{c|c|c|}
\cline{2-3}
 & Most & Least \\ \hline
\multicolumn{1}{|c|}{1} & 3 star & even pull \\ \hline
\multicolumn{1}{|c|}{2} & delicious & our condo \\ \hline
\multicolumn{1}{|c|}{3} & 3 5 & decor that \\ \hline
\multicolumn{1}{|c|}{4} & three star & 9 friend \\ \hline
\multicolumn{1}{|c|}{5} & terrible & that fine \\ \hline
\multicolumn{1}{|c|}{6} & horrible & crispy perfection \\ \hline
\multicolumn{1}{|c|}{7} & not worth & anyone because \\ \hline
\multicolumn{1}{|c|}{8} & bland & canned corn \\ \hline
\multicolumn{1}{|c|}{9} & excellent & mouth open \\ \hline
\multicolumn{1}{|c|}{10} & disappointing & rice either \\ \hline
\end{tabular}
\end{center}
\end{table}

Table VII shows the most and least discriminative words. This is defined as the standard deviation among the fitted coefficients. For example, the most discriminative phrase is ``3 star'', with fitted coefficients of (-0.525, 0.868, -0.343), unsurprisingly high for class 1 and low for the other two classes. On the other hand, all of the least discriminating words/phrases have coefficients that are essentially 0, meaning that they don't contribute much to our model prediction and could be eliminated from our feature space in model builds.

For each list, we can get almost identical results from our trained SVM model. Similar results are seen in Yu, 2017 \cite{yu2017}, which also uses SVM.

\subsection{LSTM}

Deep learning models, particularly Recurrent Neural Networks (RNN), have been proven successful in many NLP related tasks due to its nature of allowing information to persist, much like how humans think. LSTM is a special kind of RNN that solves the ``vanishing gradient'' problem when the dependency is long (Olah, 2015 \cite{olah2015}).

After experimenting with many different techniques, we find that:

\begin{itemize}
    \item Text normalization (e.g., lemmatization with part-of-speech tagging) does not improve model performance
    \item Using pre-trained word embeddings such as GLoVe (https://nlp.stanford.edu/projects/glove) helps with minimizing variance during the training phase
    \item If pre-trained word embeddings are used, it is better to not cap the vocabulary size. Otherwise, capping the vocabulary size to say 5,000 words yields better results
    \item Even though 20\% of our reviews have more than 200 tokens, while only 5\% of our reviews have more than 300 tokens, capping maximum length to 200 reduces overfitting and yields better overall performance
    \item Use dropouts in between stacked LSTM layers to reduce overfitting
\end{itemize}

\begin{table}[htbp]
\caption{LSTM Architecture Summary}
\begin{center}
\begin{tabular}{|c|c|c|}
\hline
Layer & Output Shape & Param \# \\ \hline
Embedding & (None, 200, 200) & 16,381,400 \\
LSTM & (None, 200, 64) & 67,840 \\
Dropout & (None, 200, 64) & 0 \\
LSTM & (None, 64) & 33,024 \\
Dense & (None, 64) & 4,160 \\
Dense & (None, 3) & 195 \\ \hline
\multicolumn{3}{|l|}{Total params: 16,486,619} \\ \hline
\multicolumn{3}{|l|}{Trainable params: 105,219} \\ \hline
\multicolumn{3}{|l|}{Non-trainable params: 16,381,400} \\ \hline
\end{tabular}
\end{center}
\end{table}

Table VIII shows the architecture of our final LSTM model. Since we are using the word embedding from GloVe pre-trained on Wikipedia, only a fraction of the parameters in our model are trainable, greatly reduces the computation load.

\begin{figure}[htbp]
\centerline{
\includegraphics[width=\columnwidth]{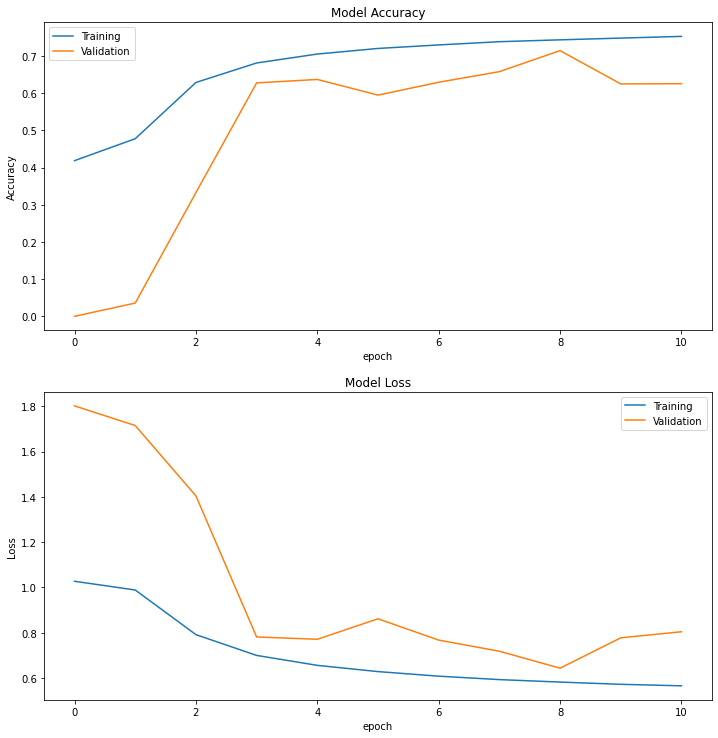}
}
\caption{LSTM Training History}
\label{fig}
\end{figure}

Fig. 11 shows the accuracy and loss per epoch during the training phase, where we use 20\% of our training set as validation set for early-stopping. As we can see, the model learns very quickly and most of the biases disappear before epoch 5.

\begin{figure}[htbp]
\centerline{
\includegraphics[scale=0.5]{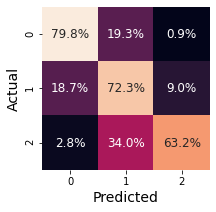}
}
\caption{LSTM Confusion Matrix}
\label{fig}
\end{figure}

Fig. 12 shows the confusion matrix for our trained LSTM model on the test set. Most of the errors come from class 2, in which the model incorrectly assigns 34\% of the reviews to class 1. This is because positive ratings tend to have shorter reviews than negative and neutral ratings -- in our dataset, only 15\% of positive ratings exceed the 200 tokens maximum length cap, while 23\% of negative and neutral ratings exceed this cap. Our experiment shows that, if we increase the maximum length cap to 300, errors become much more symmetrical. However, the model overfits and the overall performance deteriorates.

\section{Conclusion and Future Works}

In this paper, we perform an ablation study on text preprocessing techniques and compare the effectiveness of several machine learning and deep learning models on predicting user sentiments.

For machine learning models, we find that using binary bag-of-word representation, adding bi-grams, imposing minimum frequency constraints and normalizing texts have positive effects on model performance. For deep learning models, we find that using pre-trained word embeddings and capping maximum length often boost model performance.

Overall, we find simpler models such as Logistic Regression and SVM to be more effective at predicting sentiments than more complex models, in terms of both model performance and training time. More importantly, simpler models offer great interpretability, while more complex models require other techniques (e.g., LIME analysis \cite{ribeiro2016}) to understand its inner workings.

However, since deep learning models require much heavier computation than machine learning models, one possible improvement to our analysis is to run our models on better hardware setup, so we can have more relaxed constraints on maximum length and possibly use larger dataset.

Lastly, we re-emphasize that there is no one-fits-all solution to building sentiment analysis models, and the insights from this paper should only be considered as guidance. Deep learning models, albeit their lack of interpretability, tend to outperform machine learning models for more complex NLP tasks such as speech recognition, translation and automatic summarization.


\begin{thebibliography}{00}

\bibitem{sokolova2009} Sokolova, Marina, and Guy Lapalme. "A systematic analysis of performance measures for classification tasks." Information processing \& management 45.4 (2009): 427-437.

\bibitem{renault2019} Renault, T. Sentiment analysis and machine learning in finance: a comparison of methods and models on one million messages. Digit Finance (2019).

\bibitem{wang2012} Wang, Sida, and Christopher D. Manning. "Baselines and bigrams: Simple, good sentiment and topic classification." Proceedings of the 50th annual meeting of the association for computational linguistics: Short papers-volume 2. Association for Computational Linguistics, 2012.

\bibitem{saif2014} Saif, Hassan, et al. "On stopwords, filtering and data sparsity for sentiment analysis of twitter." (2014): 810-817.

\bibitem{salinca2015} Salinca, Andreea. "Business reviews classification using sentiment analysis." 2015 17th International Symposium on Symbolic and Numeric Algorithms for Scientific Computing (SYNASC). IEEE, 2015.

\bibitem{xu2014} Xu, Yun et al. “Sentiment Analysis of Yelp's Ratings Based on Text.” (2014).

\bibitem{yu2017} Yu, Boya, et al. "Identifying Restaurant Features via Sentiment Analysis on Yelp Reviews." arXiv preprint arXiv:1709.08698 (2017).

\bibitem{olah2015} Olah, Christopher. “Understanding LSTM Networks.” Understanding LSTM Networks, colah.github.io/posts/2015-08-Understanding-LSTMs/.

\bibitem{hosmer2013} Hosmer Jr, David W., Stanley Lemeshow, and Rodney X. Sturdivant. Applied logistic regression. Vol. 398. John Wiley \& Sons, 2013.

\bibitem{suykens1999} Suykens, Johan AK, and Joos Vandewalle. "Least squares support vector machine classifiers." Neural processing letters 9.3 (1999): 293-300.

\bibitem{hochreiter1997} Hochreiter, Sepp, and Jürgen Schmidhuber. "Long short-term memory." Neural computation 9.8 (1997): 1735-1780.

\bibitem{devlin2018} Devlin, Jacob, et al. "Bert: Pre-training of deep bidirectional transformers for language understanding." arXiv preprint arXiv:1810.04805 (2018).

\bibitem{chen2016} Chen, Tianqi, and Carlos Guestrin. "Xgboost: A scalable tree boosting system." Proceedings of the 22nd acm sigkdd international conference on knowledge discovery and data mining. 2016.

\bibitem{ribeiro2016} Ribeiro, Marco Tulio, Sameer Singh, and Carlos Guestrin. ````Why should i trust you?'' Explaining the predictions of any classifier.'' Proceedings of the 22nd ACM SIGKDD international conference on knowledge discovery and data mining. 2016.

\bibitem{dasilva2014} Da Silva, Nadia FF, Eduardo R. Hruschka, and Estevam R. Hruschka Jr. "Tweet sentiment analysis with classifier ensembles." Decision Support Systems 66 (2014): 170-179.

\end{thebibliography}
\end{document}